\documentclass[conference]{IEEEtran}
\usepackage{times}
\usepackage{todonotes}
\usepackage[numbers]{natbib}
\usepackage{multicol}
\usepackage{graphics,graphicx,caption,float,subcaption,booktabs,xcolor,multirow,array,color,ifthen,tabu,colortbl,dblfloatfix,url,xparse,mathtools,patchcmd,algorithm,algorithmic,amssymb,xspace,nicefrac,microtype,amsmath,amsfonts,bm,ragged2e,tikz,stackengine,etoolbox,xpatch,enumerate,xstring,setspace,tabularx,makecell,changepage,cuted,titlesec,wrapfig,tcolorbox, sidecap,comment, bbding, placeins}
\usepackage{gensymb,comment}
\usepackage[pagebackref=true,breaklinks=true,colorlinks=true,bookmarks=false,citecolor=blue]{hyperref}
\hypersetup{
colorlinks=true,
linkcolor=blue,
filecolor=magenta,      
citecolor=blue
}

\usepackage{siunitx}
\newcommand{\e}[1]{\times 10^{#1}}
\title{\LARGE \bf LEAP Hand: Low-Cost, Efficient, and Anthropomorphic Hand\\for Robot Learning}

\author{Kenneth Shaw, Ananye Agarwal, Deepak Pathak \\ Carnegie Mellon University}

\begin{document}
\newcommand{\our} {LEAP Hand\xspace}
\newcommand{\ourablation}{LEAP-C Hand\xspace}
\newcommand{\ourshort}{LEAP\xspace}
\def\@onedot{\ifx\@let@token.\else.\null\fi\xspace}
\DeclareRobustCommand\onedot{\futurelet\@let@token\@onedot}
\newcommand{\figref}[1]{Fig\onedot~\ref{#1}}
\def\etal{\emph{et al}\onedot}
\newcommand{\secref}[1]{Sec\onedot~\ref{#1}}
\newcommand{\tabref}[1]{Tab\onedot~\ref{#1}}
\newcommand\ananye[1]{\textcolor{red}{#1}}
\makeatletter
\let\@oldmaketitle\@maketitle
\renewcommand{\@maketitle}{\@oldmaketitle
\includegraphics[width=\linewidth]{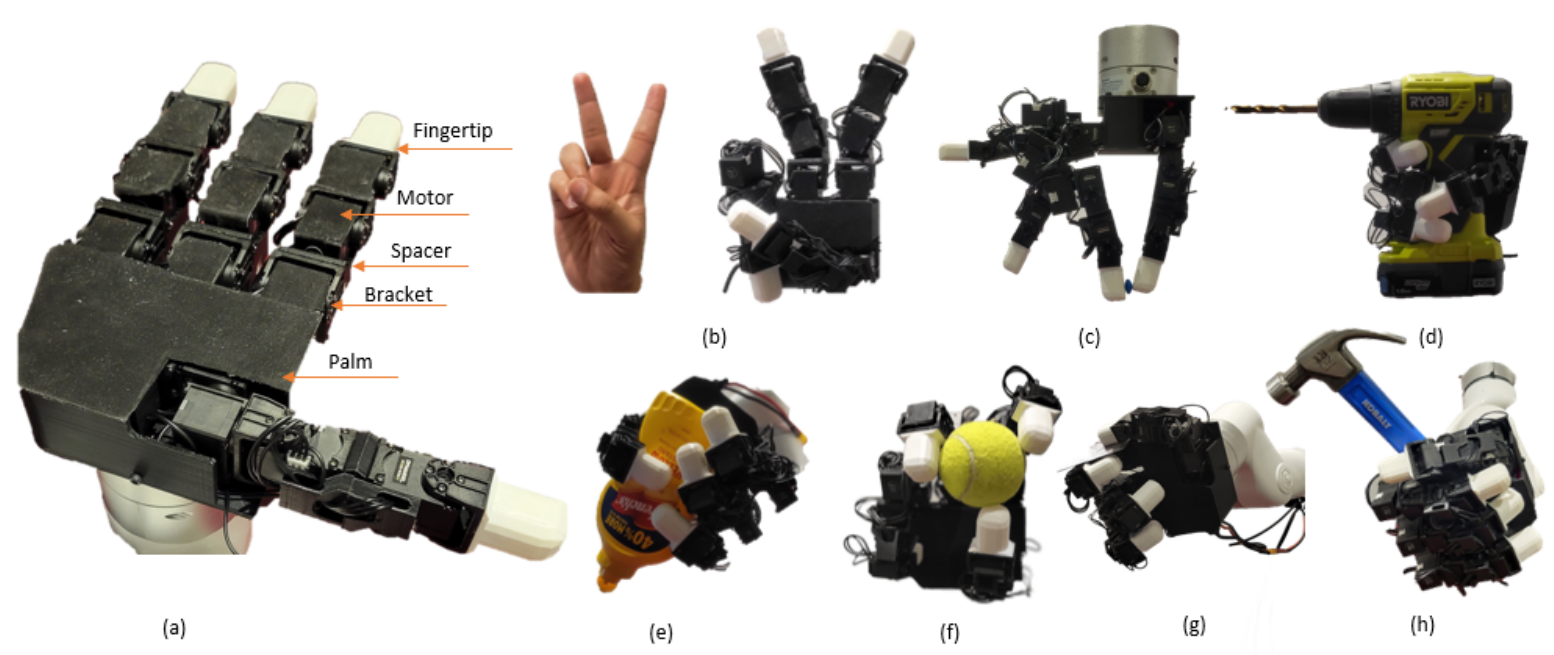}
  \centering
  \vspace{-0.2in}
  \captionof{figure}{\small (a)  \our is an anthropomorphic dexterous robot hand designed for robot learning research. It can be assembled in under 4 hours for 2000 USD, is composed of readily available parts, and is robust. (b) to-scale comparison of \our and a human hand (c-h) \our in different power and precision grasps holding common objects. The hand design and code will be open-sourced to democratize access to hardware for anthropomorhic dexterous manipulation. Video, assembly instructions, and sim2real pipeline at~\url{https://leap-hand.github.io/}}
  \label{fig:teaser}
  \vspace{-0.1in}
  \bigskip}

\makeatother
\maketitle
 \thispagestyle{empty}
\pagestyle{empty}

\begin{abstract}

Dexterous manipulation has been a long-standing challenge in robotics. While machine learning techniques have shown some promise, results have largely been currently limited to simulation.  This can be mostly attributed to the lack of suitable hardware. In this paper, we present \our, a low-cost dexterous and anthropomorphic hand for machine learning research. In contrast to previous hands, \our has a novel kinematic structure that allows maximal dexterity regardless of finger pose. \our is low-cost and can be assembled in 4 hours at a cost of 2000 USD from readily available parts.  It is capable of consistently exerting large torques over long durations of time. We show that \our can be used to perform several manipulation tasks in the real world---from visual teleoperation to learning from passive video data and sim2real. \our significantly outperforms its closest competitor Allegro Hand in all our experiments while being 1/8th of the cost.  We release detailed assembly instructions, the Sim2Real pipeline and a development platform with useful APIs on our website at ~\url{https://leap-hand.github.io/}
\end{abstract}

\section{Introduction}
Hand dexterity has been critically responsible for human cognition through active manipulation, tool use, and governing how humans learn from the world~\cite{libertus2016motor, bruce2012learning, gibson1988exploratory}.
Replicating the dexterity of the human hand with a robot hand has been a long-standing challenge in robotics. Machine learning techniques have recently shown promise in areas such as learning from humans. However, unlike the learning successes in locomotion ~\cite{miki2022learning,agarwal2022legged} across truly diverse terrains, robotic manipulation results in the real world have mostly been limited to one degree-of-freedom parallel jaw grippers~\cite{Bhardwaj-CoRL-21,NEURIPS_rb2,sundermeyer2021contact}.  In contrast, dexterous manipulation has largely been limited to simulation~\cite{chen2021system, huang2021geometry} with comparatively fewer real-world results ~\cite{andrychowicz2020learning, sievers2022learning, morgan2022complex, handa2022dextreme, chen2022visual}.

\begin{figure*}[t!]
 \centering
\includegraphics[width=0.95\textwidth]{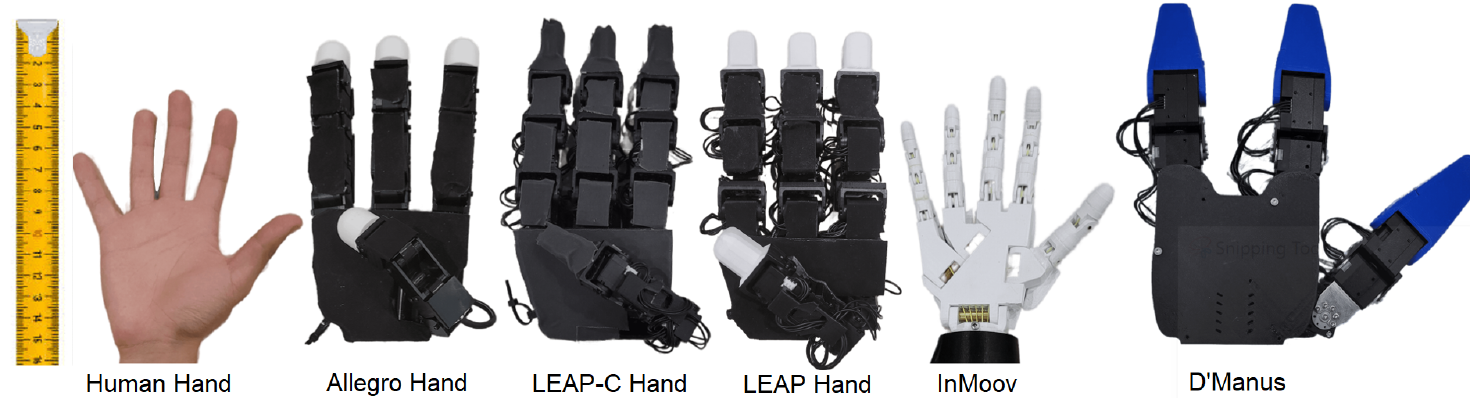}
\caption{\small \textbf{Relative size of popular robot hands to scale.} \textit{Left to right,} adult human hand,  Allegro Hand \cite{allegro}, \ourablation, \our, Inmoov \cite{inmoov}, D'Manus \cite{bhirangi2022all}. \our is similar in size to Allegro and $\sim30\%$ larger than a human hand. D'Manus is considerably larger than the rest.  Because of the tendon-driven nature, Inmoov is the smallest robotic hand. The hands are accurate to scale.}
\vspace{-0.05in}
 \label{fig:comparison}
\end{figure*}

A major bottleneck in democratizing dexterous manipulation has been the hardware. Tendon-based hands like Shadow~\cite{ShadowHand}, while impressively capable~\cite{andrychowicz2020learning}, cost over 100K USD and often require significant maintenance \cite{akkaya2019solving} due their complicated nature. While Inmoov \cite{inmoov} is inexpensive and open source, it only has 5 actuators on weak tendons. Therefore, direct-driven hands have been the popular alternative for many applications~\cite{lee2016kitech, allegro}.  The Allegro Hand has been a popular direct-driven hand, but it is often unreliable, difficult to repair, and does not have an anthropomorphic kinematic structure, (see Fig \ref{fig:comparison}) and is expensive at over \$16K.  Please see Section \ref{sec:related} for further analysis.

As a result, only a few labs have access to hardware capable of complex dexterous tasks. This is in stark contrast to 2-finger grasping or locomotion \cite{xarm, franka, a1, go1} where readily available hardware allows results to be easily reproduced and improved upon by the community. Following this analogy, good hand hardware for machine learning must be durable, repeatable, low-cost, versatile, and ideally anthropomorphic to enable easy transfer learning from humans.

We propose \our -- a dexterous, extremely \textit{low-cost} and \textit{robust} hand for robot learning, built from off-the-shelf or 3D-printed parts. Our hand can be assembled in under 4 hours at a cost of 2000 USD, which is 1/8th the cost of the Allegro Hand and 1/50th to that of ShadowHand. While we acknowledge this is still not affordable for all, we believe it is a step towards democratizing dexterous manipulation research. We show through a number of rigorous experiments that \our is both robust, durable, and able to exert large torques over long periods of time. 
 Additionally, our robot hand it is easily repairable in-house just using a standard \$250 3D printer and does not need to be sent out for repair.

Although robustness and low-cost is critical, they should not come at the cost of dexterity and anthropomorphism. We believe a good versatile hand is the one that is both \textit{dexterous} as well as \textit{anthropomorphic} because much of the world around us, for instance, doors, kitchens, tools, or instruments, are designed with human hands in mind, making it easier to learn by watching humans act. In \our, we aim to maximize dexterity while being kinematically similar to a human hand.

\begin{figure*}[t!]
 \centering
\includegraphics[width=\linewidth]{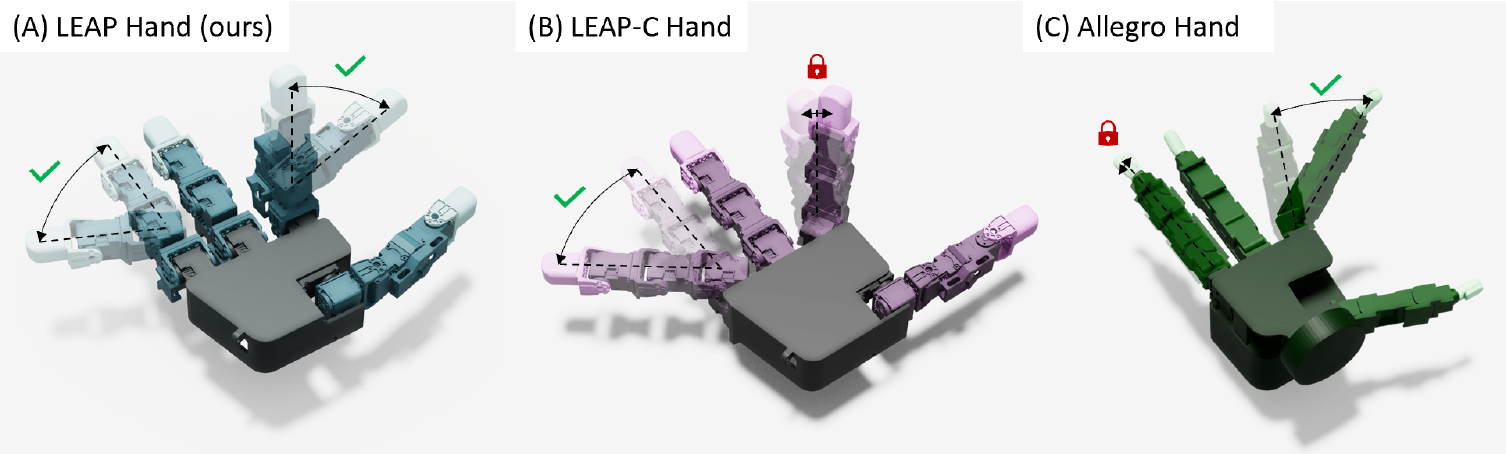}
\caption{\small Comparison of MCP joints in different robot hands and their dexterity and two different positions. (A) In \ourablation there is a large range of motion at extended but not flexed position (B) In \our, at flexed and extended positions, the fingertip has a large range of motion. (C) In Allegro, there is a large of motion at flexed but not extended position.} 
\vspace{-0.15in}
 \label{fig:kinematics}
\end{figure*}

Since the ball joint at the human knuckle (Metacarpophalangeal aka MCP) cannot be replicated with direct-driven hands, it must instead be approximated using two separate motors. Prior work in direct driven hands has converged primarily to two designs, see Fig~\ref{fig:kinematics}, one that allows abduction-adduction of fingers in open hand pose and the other that only allows with the finger flexed upwards. However, both of these lose one degree of freedom (DoF)---either in the flexed or extended position of the finger. In \our, we propose a new kinematic mechanism to facilitate \textit{universal abduction-adduction} for direct-driven hands that retain all degrees of freedom in all finger positions. We demonstrate that this leads to higher dexterity and for improved grasping and in-hand manipulation. 

Finally, we show that \our easily integrates with existing results in robot learning. For instance, YouTube video-based learned teleoperation and behavior cloning. In addition to the physical robot hardware, we also release an Isaac Gym-compatible simulator for the \our and show sim2real transfer for a contact-rich task of blind in-hand rotation of a cube~\cite{ma2011dexterity}. This shows that the hardware and simulation are accurate and that complex tasks trained in simulation can be transferred to the real hand. We \textbf{open source} the URDF model, assembly instructions, ROS/Python API, mapping methods from human hands to \our, and an Isaac Gym simulation environment at~\url{https://leap-hand.github.io/}.

\section{Related Work}
\label{sec:related}
\vspace{0.05in}
\textbf{Robot Hands}$\quad$
Shadow~\cite{ShadowHand} and ADROIT~\cite{kumar2014real} hands paved the way to enable complex, contact-rich dexterous tasks with an anthropomorphic ball joint MCP.  \cite{andrychowicz2020learning,akkaya2019solving}. However, they are costly (100k USD) and require constant maintenance.
In contrast, the Inmoov hand is 3D printable, tendon-driven, and human-like\cite{inmoov}. It has only one DoF per finger and is reliant on tendon actuation which is difficult to calibrate and can be inaccurate.  Bauer et. al. \cite{bauer2022towards} present a soft tendon-driven hand that is very flexible for many configurations.  Unfortunately, it is difficult to simulate due to its deformable nature \cite{sofa, duriez2017soft}. In contrast to tendon-driven hands, which have motors in the wrist, the Allegro Hand~\cite{allegro} has its motors in the finger joints. It is most popular in research labs~\cite{handa2020dexpilot,sundaralingam2019relaxed, sivakumar2022robotic, arunachalam2022dexterous, qin2022from}  because it is relatively cheap (16k USD). However, users find the motors in the fingers to be weak for many everyday tasks. Moreover, the closed-source components are difficult to repair or replace.  Additionally, its kinematic structure is not anthropomorphic or dexterous as we demonstrate.  The ROBEL suite (which includes D'Manus) ~\cite{bhirangi2022all} is more robust, open-sourced, and easy to build. However, it has only two fingers and a thumb---making it significantly different from a human hand. 
Yuan et. al. \cite{yuan2020design} accomplish within-hand manipulation using rollers attached to the fingers. Humans lack this degree of freedom and manipulate objects in a very different manner. 
\cite{clone, kim2019fluid, kim2021integrated} have shown impressive results and hold a lot of promise by using fluids and linear actuators to move the fingers, but these hands are not readily available and too complicated to quickly produce, use, and maintain for robot learning.

\vspace{0.05in}\textbf{Rapid Manufacturing}$\quad$
Aluminum machining is traditionally used to create strong parts but is prohibitively difficult and expensive.  Manufacturing plastic parts includes a cumbersome process of mold making, casting, curing, and support removal~\cite{pneuflex}.  In contrast, additive manufacturing can be used to create parts very quickly for prototyping. In our paper, we leverage recent advancements in extruders, hot-ends, and motors made by the open-source Reprap Community~\cite{reprap}.  This allows us to directly print soft flexible filaments like Ninjaflex~\cite{Ninjatek}.  We use this to create many parts for \our like the hard palm and soft rubber fingertips.

\textbf{Learning Dexterity}
Using a Shadow hand and Sim2real, Andrychowicz et. al.~\cite{andrychowicz2020learning, akkaya2019solving,kumar2021rma} accomplish in-hand rotation for a variety of objects. Policies that scale to thousands of objects can also be trained in simulation~\cite{chen2021system,huang2021generalization}.  \cite{nair2020awac} uses the D'Hand to reposition a valve. In-hand rotation of Baoding Balls using the Shadow Hand trained purely in the real-world~\cite{nagabandi2020deep}, and pipe insertion using the D'Hand~\cite{gupta2021reset} are other notable examples of dexterous manipulation.

Several recent works focus on supervising policies of robot hands~\cite{wang2020rgb2hands, hmr, feng2021collaborative, FrankMocap_2021_ICCV}. from MANO~\cite{MANO:SIGGRAPHASIA:2017} parameters which parameterize a human hand.  Closely related is the teleoperation of robot hands from real-time video \cite{handa2020dexpilot, robo-telekinesis}, which can be used to guide learning and improve sample-efficiency \cite{qin2021dexmv, rajeswaran2017learning}.   Hand poses can be extracted from video data available on the web to learn manipulation policies~\cite{qin2021dexmv, mandikal2021dexvip}. Large-scale pre-training using internet videos is helpful for efficiently training robot hands for downstream tasks using a few task specific-demos \cite{videodex} and also on non-dexterous manipulation \cite{bahl2022human, pari2021surprising}.

\section{Kinematic Design and Analysis}
The kinematic structure of a hand refers to the arrangement of its joints which determines the different poses and forces of motion it can apply.  First, \our should be as anthropomorphic as possible so that data from humans can be used to learn skills~\cite{bahl2022human, videodex, qin2021dexmv} with machine learning. This can be done using methods such as teleoperation with VR gloves or extracting keypoints from videos of human hands~\cite{robo-telekinesis}. In addition, \our should be dexterous for tasks such as in-hand manipulation from sim2real.  In this section, we propose a robot hand design that is both anthropomorphic and dexterous.

\begin{figure*}[b!]
 \centering
 \vspace{-0.1in}
\includegraphics[width=\linewidth]{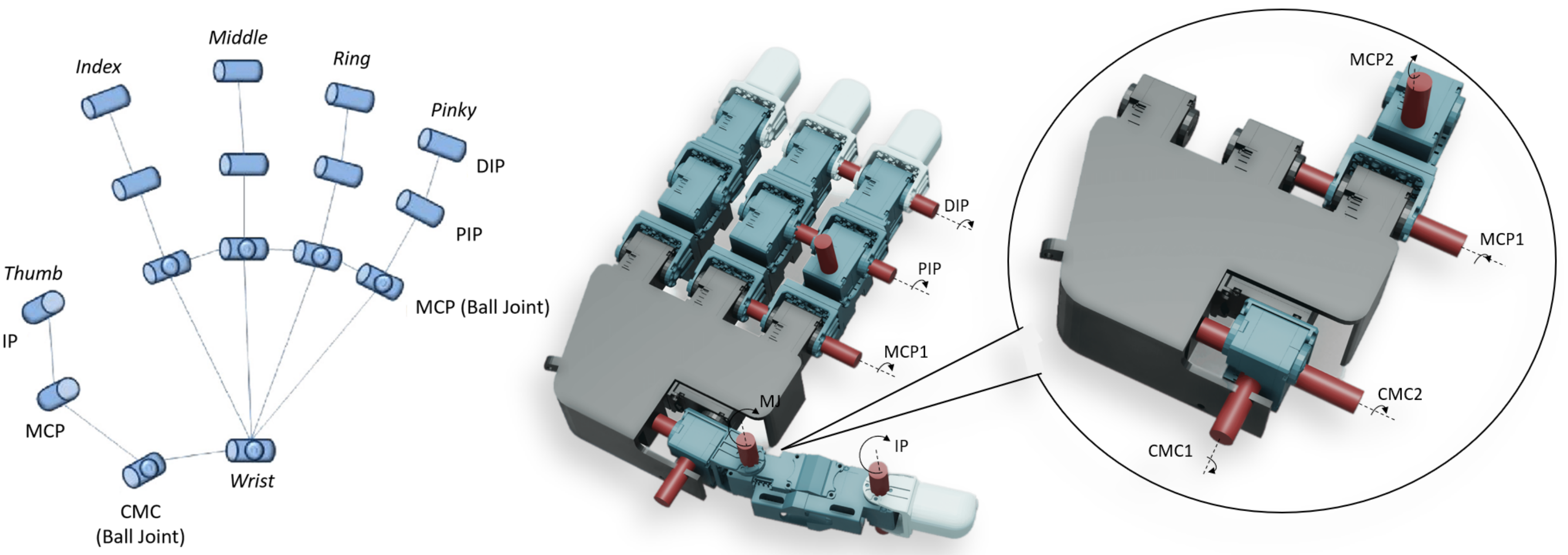}
\caption{\small The human hand kinematics above has ball joints at the MCP and CMC joints.  These are difficult joints for low-cost hands to include. Left Figure from \cite{cerulo2017teleoperation}. Comparison of MCP joints in different robot hands. (A) In \ourablation there is a large range of motion at extended but not flexed position (B) In \our, at flexed and extended positions, the fingertip has a large range of motion. (C) In Allegro, there is a large of motion at flexed but not extended position.} 
 \label{fig:kinematics_human}
\end{figure*}

In the human hand, there are four main degrees of freedom in each finger (Fig. \ref{fig:kinematics_human}).  The knuckle or metacarpophalangeal (MCP) is a ball joint with two degrees of freedom that allows abduction/adduction and flexion/extension. The joint closest to the knuckle is called the proximal interphalangeal (PIP) joint. The last joint, closest to the fingertip, is the distal interphalangeal (DIP). The PIP and DIP are hinge joints, each with one degree of freedom.  The human hand features an opposable thumb which allows the application of force in opposition to other fingers. This enables a variety of power and precision grasps~\cite{liu2014taxonomy}.  To easily map motions, a robot hand must have analogous joints to a human hand.  

To replicate this structure, it is alluring to use tendons like robot hands such as the  ShadowHand~\cite{ShadowHand}. Such tendon-driven hands can store the large motors needed to drive them in the wrist, enabling greater flexibility in joint design and introduce ball joints. However, they are very expensive (100K USD), complicated or hard to maintain. As a result, cheaper direct-driven alternatives~\cite{allegro, lee2016kitech} have been more popular. 

\begin{figure}[t]
\centering
\includegraphics[width=\linewidth]{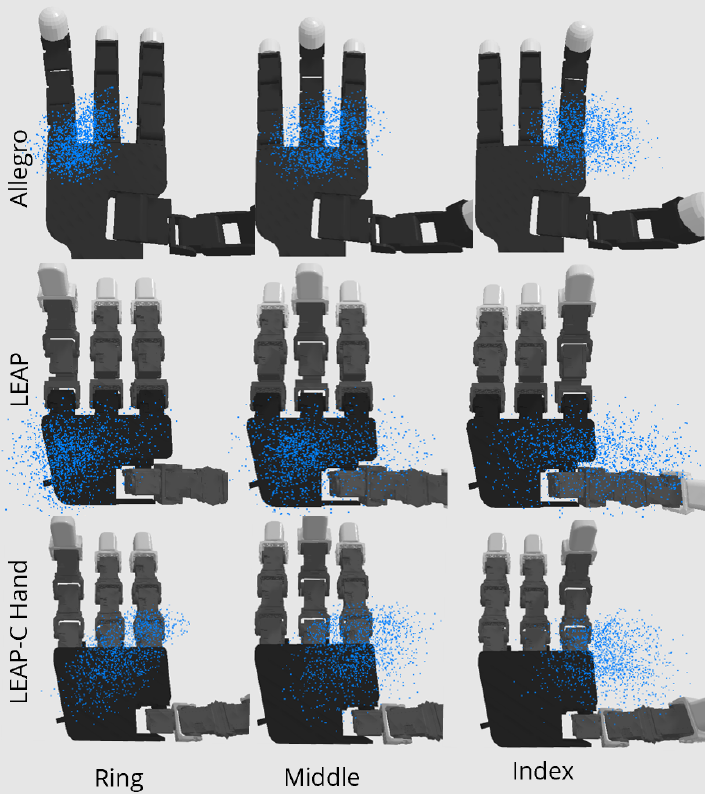}
\caption{\small We compare the possible positions of opposability of the thumb and each of the other fingers on each of the three hands.  We find that \our has the best even spread on top of the palm and a very large contact area.}
\vspace{-0.25in}
\label{fig:kinematic_comparison}
\end{figure}

\subsection{Universal Abduction-Adduction Mechanism}
Direct-driven hands must store the motors inside the fingers so they are limited in kinematic structure and cannot precisely imitate the human hand. Since the PIP and DIP joints are hinge joints, they are easily modeled, each with a single actuator. A ball joint cannot be modeled in this way and is typically approximated using two motors (MCP-1, MCP-2) arranged close together~\cite{lee2016kitech}. Prior seminal work has proposed two designs for this (Fig.~\ref{fig:kinematics}). However, both of these designs, Allegro and \ourablation, lose one degree of freedom in either the extended or closed position. As a result, Allegro is less dexterous when extended whereas \ourablation (like C-Hand in~\cite{lee2016kitech}) is less dexterous when closed.  

The reason for the lost dexterity in both \ourablation and Allegro is that the axis of the motor responsible for adduction-abduction (MCP-2) is fixed to the palm of the hand.  In \ourablation, the axis is perpendicular to the plane of the palm, whereas, in Allegro, it lies in the plane of the palm. Thus, when the finger becomes parallel to this axis, that DoF is ineffective.  Please see Figure \ref{fig:kinematics} and the kinematic tree in the supplemental.

In \our, we propose a new \textbf{\textit{universal abduction-adduction mechanism}} for the fingers such that they can retain all degrees of freedom at all MCP positions. Instead of the MCP-2 axis being fixed to the palm (i.e., of motor responsible for adduction-abduction), the key idea is to bring the axis to the frame of reference of the first \emph{finger} joint and arrange it such that it is always perpendicular to it. This allows the finger to have adduction-abduction in all positions (Fig.~\ref{fig:kinematics}). Thus, \our has adduction-abduction in the extended position (similar to \ourablation) as well as pronation/supination in the flexed position (similar to Allegro). 

\subsection{Evaluating Manipulability via Thumb Opposability}
Chalon et.al.~\cite{chalon2010thumb} and Lee et.al.~\cite{lee2016kitech}, have shown that what makes a hand more versatile is not merely the degree of abduction-adduction but also its thumb opposability volume. We test our design against Allegro and \ourablation, a baseline hand we manufacture with the same motors and parts as \our. In Fig.~\ref{fig:kinematic_comparison}, we plot the intersection of the thumb and finger workspaces for each hand and compute a thumb opposability metric~\cite{chalon2010thumb}.  In Table \ref{tab:opposability}, we show that \our combined with the new MCP joints in the secondary fingers is better placed and is more dexterous compared to other available hands because of the increased opposable volume. 

Next, manipulability measures the ease with which the fingertip can be moved in various directions at a particular joint pose.  We use metrics introduced by Yoshikawa et. al. ~\cite{yoshikawa1985manipulability}. To evaluate this, many calculate the manipulability ellipsoid from the end-effector Jacobian which models the directions in which the end-effector can move.  We compute the volume of this ellipsoid using the following equation:

$$
w=\sqrt{\operatorname{det}\left(\boldsymbol{J}(\boldsymbol{q}) \boldsymbol{J}(\boldsymbol{q})^{\mathrm{T}}\right)}
$$

where $\boldsymbol{q}$ is the joint configuration and $\boldsymbol{J}$ is the Jacobian of the end-effector.  Note that because we are calculating volumes, a hand that can only move in one or two cartesian directions will have a volume close to zero at that pose.  In three key poses, we show that \our has consistently larger ellipsoids of greater volume for both the cartesian and angular components of the jacobian.  This means that \our has better movement at the fingertips in these few poses and a higher manipulability metric leading to more dexterity (Table~\ref{tab:opposability}).

\begin{table}[t]
\centering
\resizebox{0.95\linewidth}{!}{
\begin{tabular}{lccc}
\toprule
Robot/Position & Down (m$^3$) & Up (m$^3$) & Curled (m$^3$) \\
\midrule
\multicolumn{4}{l}{\textit{Allegro Hand}} \\
Linear & $8.11\e{-9}$ & $3.98\e{-13}$ & $2.39\e{-5}$ \\
Angular & 0 & 0 & 0\\
\midrule
\multicolumn{4}{l}{\textit{\ourablation}} \\
Linear & $1.60\e{-12}$ & $1.23\e{-10}$ & $9.28\e{-5}$  \\
Angular & $1.02\e{-13}$ &	$1.02\e{-9}$ & $2.02\e{-13}$\\
\midrule
\multicolumn{4}{l}{\textit{\our (ours)}} \\
Linear & $2.02\e{-6}$ & $2.42\e{-6}$ &$4.51\e{-5}$  \\
Angular & $1.20\e{-5}$ & $1.20\e{-5}$& $1.20\e{-5}$\\
\bottomrule
\end{tabular}}
\caption{\small  We show the manipulability ellipsoid volume for both the linear and angular component at three different finger positions, down, all the way up, and then halfway/curled.  We find that \our has a large manipulability ellipsoid at all three configurations.}
\vspace{-0.07in}
\label{tab:opposability}
\end{table}

Finally, we show that increased dexterity leads to practical benefits as well. In the grasping test (Sec.~\ref{sec:grasping}), we find that \our is able to grasp more objects tightly. In the blind in-hand cube rotation task (Sec.~\ref{section:inhand}), we find that \our is able to rotate the cube much faster than Allegro. 

\section{Hand Design Principles}
A good kinematic design must be realized effectively in hardware. In particular, the hardware should be low-cost, easy to repair, and robust. 

\subsection{\textbf{Low-cost and Easy to Repair}}
In contrast to locomotion or manipulation with two-fingered grippers, real-world research in dexterous manipulation has been limited. This can be attributed in large part to the lack of suitable dexterous hand hardware. Commonly used dexterous hands such as ShadowHand~\cite{ShadowHand} and AllegroHand~\cite{allegro} cost 100K and 16K USD, respectively, and must be sent back to manufacturers for repair in case of damage. This hardware is out-of-budget or impossible to maintain for many researchers allowing only a small fraction to work on real-world dexterous manipulation. On the other hand, due to the availability of cheap and reliable locomotion \cite{a1, go1} and manipulation \cite{xarm, franka, ur5} hardware, a large community of researchers is able to build off of each others' work and drive progress. 

A suitable hand should therefore be as accessible as possible. This implies that it should be low-cost and easy to repair. In \our, we accomplish this by using as many off-the-shelf parts as possible and fabricating the rest using only a commodity 3D printer that costs around 200 USD. It can be assembled in under 4 hours.

\our is designed to be modular.  This allows key features of the robot hand to be changed, such as the length or number of fingers and the distances between each of the fingers in the palm for particular learning tasks or for analysis.  Additionally, the modularity makes the hand easily repairable with only a few distinct parts.

\begin{table}[t]
\centering
\resizebox{0.95\linewidth}{!}{
\begin{tabular}{lccc}
\toprule
Opposability Vol. & Index (mm$^3$) & Middle (mm$^3$) & Ring (mm$^3$) \\
\midrule
Allegro & 409,135 & 348,809 & 204,281  \\
\ourablation & 834,516 & 743,764 & 638,605\\
\our (ours)         & \textbf{1,125,556} & \textbf{1,056,746} & \textbf{804,618} \\
\bottomrule
\end{tabular}}
\caption{\small  We show the finger-to-thumb opposability volume in $mm^3$ by randomly sampling 25,000 joint configurations and finding the instances at which both fingers touch and recording that contact point. The volume of this area of contact is calculated and reported.}
\vspace{-0.07in}
\label{tab:opposability}
\end{table}

\begin{figure}[b]
\centering
\includegraphics[width=\linewidth]{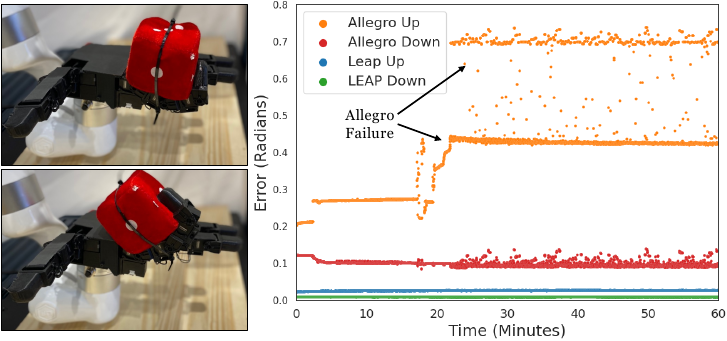}
\vspace{-0.22in}
\caption{\small \textbf{Repeatability test}. \textit{Left:} An illustration of \ourablation performing repeating grasp-ungrasp on a small plush dice using one joint for an hour. \textit{Right:} Comparison of \ourablation and Allegro~\cite{allegro}. After just 15 minutes, the Allegro Hand cannot maintain movement. In contrast, \our continues to maintain movement with minimal joint error ($<0.05$ rad). Videos at~\url{https://leap-hand.github.io/}}
\label{fig:repeatability}
\end{figure}

\subsection{\textbf{Robustness}}
Learning, whether via teleoperation, behavior cloning, or reinforcement learning, on a robot hand can be notoriously harsh on hardware, especially when it is placed on a robot arm~\cite{handa2020dexpilot,sivakumar2022robotic}. Due to the movement of the arm, the hand may repeatedly collide with the table and objects it is trying to grasp. A robot hand should be robust to such treatment and continue to function reliably without breaking. In addition, a robot hand must be able to impart large torques. This is required for lifting heavy objects or using heavy tools like drills or hammers. 

While 3D printing is fast and inexpensive, the resulting parts are often not strong enough. One alternative is custom metal machined parts. However, we avoid these as they add significant cost and require specialized skill and equipment to manufacture. We instead rely on inexpensive (\$10) off-the-shelf professionally extruded reinforced plastic brackets from Robotis \cite{robotis} that are designed to withstand wear and tear.  We only print the palm and smaller wire guide spacers using a commercial 3D printer.

The joints in \our are designed to exceed the strength of the human hand. We choose motors geared to high torque output for their size while still being capable of a hand-like joint movement velocity of around 8 rad/sec. The amount of motor mass inside the hand is maximized compared to the size of the hand, and every other component is minimized. This enables the hand to be as strong as possible for its human-like form factor.  Because these motors are so powerful, we support current- or torque-limiting them as in Section~\ref{sec:fab} to manipulate fragile objects and increase the durability of the hand.

\begin{figure}[b]
 \centering
 \vspace{-0.1in}
 \includegraphics[width=0.95\linewidth]{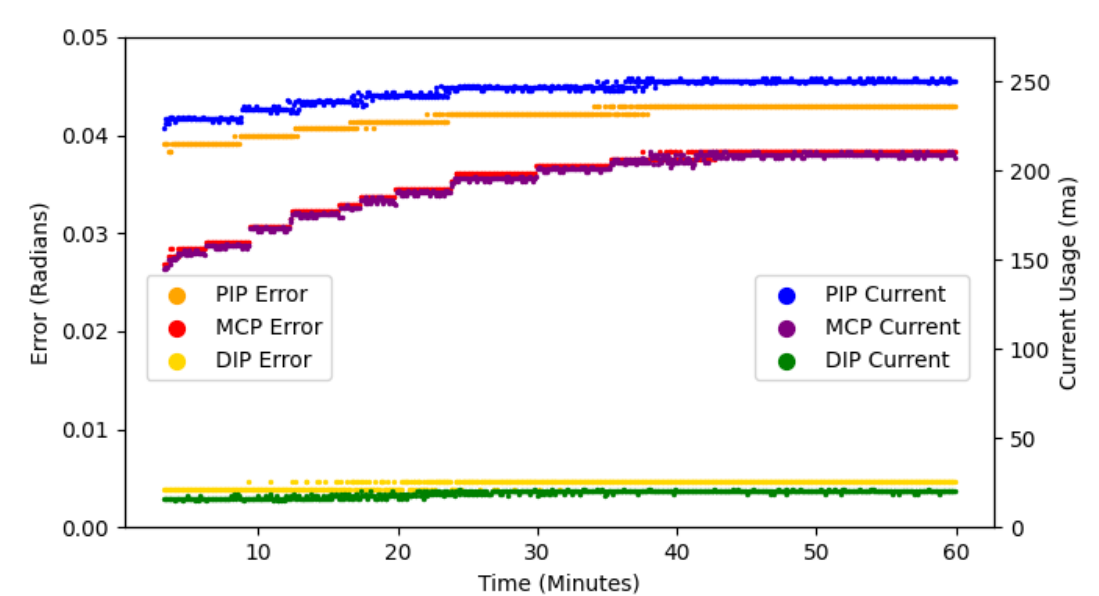}
 \vspace{-0.1in}
\caption{\small \textbf{Endurance Test}. We balance a heavy 2kg weight on only one fingertip of \our for one hour.  On the left axis, we show that the angle error of commanded vs actual remains small.  The right axis shows that the current use does initially increase with the temperature of the motor. However, it still withholds the weight and uses less than half of its maximum possible current of 600ma.}
 \label{fig:drill_hold}
\end{figure}

\subsubsection{Endurance test}
To test the strength of the hand over a long period of time, we hang a 2kg weight on one of the fingertips.  This pose is similar to if a person was holding a half gallon of milk up with one finger without using their palm as support.  We find that \our is able to continuously hold the grasp for an hour with only a small angle error.  While the current usage gradually increases initially, it stabilizes along with the temperature of the motors, which remain cool. The current usage of the top motor reaches ~250mA, which is still less than half of the maximum possible.  The Allegro Hand is not powerful enough to complete this test.  See Figure \ref{fig:drill_hold} for a plot of the results.

\subsubsection{Repeatability test}
We test the consistency and accuracy of \our against Allegro Hand by running them continuously for 1 hour in a grasping scenario as in Figure \ref{fig:repeatability}. We continuously raise and lower a small (25g) plush dice strapped onto the finger by commanding one of the base finger joints up and down at 5Hz.  The error of the desired joint angle compared to the actual joint angle is graphed through time.  

 \our has a consistent error of 0.025 radians in the up position and 0.005 in the down position (Fig. ~\ref{fig:repeatability}), which is reasonable given the PID controller and the 750mA current limit.  On the other hand, the Allegro hand starts at a much higher error. After 15 minutes, it begins to fail and then completely fails to move on one out of three grasps.  This was not a failure of the position sensor in the motor. The strain of the continuous grasping on the motor caused overheating such that the motor was not able to apply required torques.

\begin{table}[t]
\begin{center}
\footnotesize
\setlength{\tabcolsep}{3pt}
\begin{tabular}{l|c|c}
\toprule
\textbf{Hand} &  \textbf{Strength $(N)$} &  \textbf{Power Density \scriptsize{$N \times DOF/(cm^2$)}}\\
\midrule
Bauer et. al~\cite{bauer2022towards} & 37.4  & 0.677 \\
Allegro Hand~\cite{allegro}  & 8.5 & 0.35 \\
D'Manus~\cite{bhirangi2022all} & 27.8 &  0.313\\
Inmoov Hand~\cite{inmoov} & 5.8 & 0.116\\
Adult Human Hand & 26.5 & 2.199\\
\midrule
\our  & 19.5 & \textbf{1.045}  \\
\ourablation & 21.5 & \textbf{1.15}\\
\bottomrule
\end{tabular}
\caption{\small \textbf{Pullout Test}. A resistance comparison of each hand to pullout force which correlates to grasping strength.  Power density is the total amount of motor force per square area of the hand.}
\label{tab:pullout}
\end{center}
\vspace{-0.2in}
\end{table}
\vspace{-0.0in}
\subsubsection{Pull-out force test}
\label{sec:strength}
\begin{wrapfigure}{r}
 {0.25\textwidth}
 \vspace{-0.05in}
\begin{center}
\frame{\includegraphics[width=0.23\textwidth]{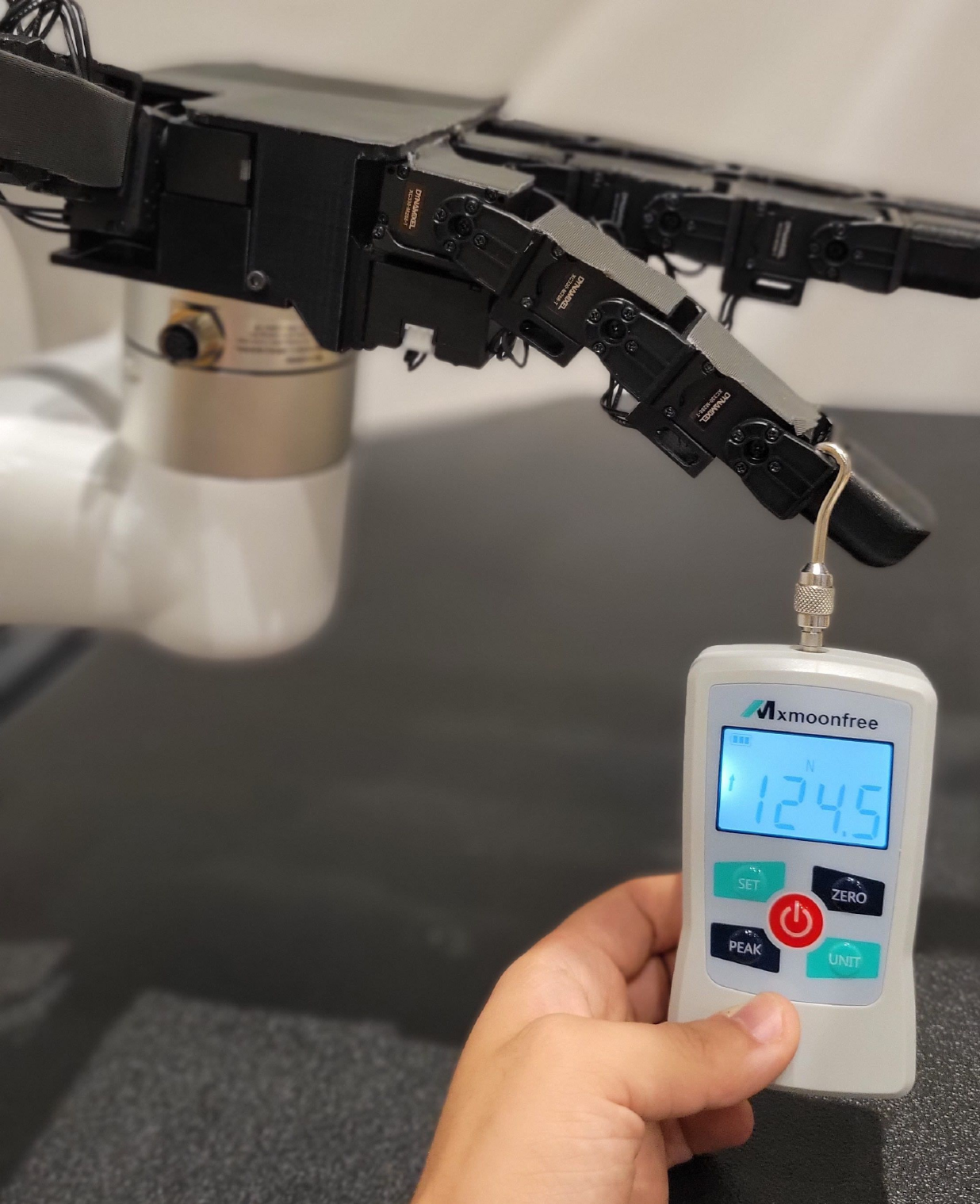}}
\caption{\small \textbf{Pullout Test}. A pullout force is applied and the maximum force is recorded before the hand has a $15\degree$ error or slipping.}
\label{fig:force}
\end{center}
\vspace{0.0in}
\end{wrapfigure}

This test measures the amount of momentary outward force that can be resisted by a flexed finger from a hooked force gauge before failure. Failure is defined as a motor or gear slipping or a finger deviating more than 15 degrees from its commanded position (Fig.\ref{fig:force}). The force returned correlates with the grip strength.

Tab.~\ref{tab:pullout} compares force for robot and human hands. D'Manus~\cite{bhirangi2022all} is the strongest due to its large motors. Of the anthropomorphic hands, \our performs the best, exceeding the grip strength of a human. The Allegro Hand is weak because of smaller motors explaining why it struggles in many grasping tasks. The tendon-driven hands, Bauer et. al, and Inmoov do not perform that much better in this test even though they can store large motors in their wrists and arms.  We find that these tendons often slip and cannot provide that much force at the end-effector.

\section{Fabrication and Software}

\noindent\textbf{Fabrication}$\quad$ First, each of the 3D printed components must be fabricated ( Fig.~\ref{fig:teaser}).  A \$200 Ender 3 3D printer \cite{ender} was used with PLA plastic over a 2 day period, but any consumer-grade FDM printer will suffice.  Each of the two palm pieces is printed along with fingertips and finger spacers.  We collect the 3D printed parts, plastic extruded brackets, the Dynamixel motors~\cite{dynamixel}, U2D2 controller, and assorted cabling. The fingers are assembled individually using brackets, 3D-printed finger spacers, and motors. The assembly process for \our takes around 4 hours. Then each of the fingers is mounted onto the palm, and their firmware is flashed for control. The hand interfaces with the computer using a USB cable and ROS, Python, or C++. The 4-finger \our weighs 595g and can be easily mounted to a variety of robot arms.  Full video instructions of the assembly process is on our website.
\label{sec:fab}

\vspace{0.05in}\noindent\textbf{Software}$\quad$
A variety of control modes are supported on \our: position control, current control, current-based position control, and velocity control. Position control enables the hand to create torques to match a desired position on the motors which is typical of many PID-based controllers. Current control mode enables a desired torque to be applied to the motors. Current-based position control mode enables PID-based position control but also caps the maximum current and torque. This enables the hand to follow position commands but also prevents large torques, which can be unsafe for the robot and the environment around it.

\noindent\textbf{Simulation}$\quad$
We construct a detailed 3D assembly of the hand as used in (Fig.~\ref{fig:kinematic_comparison}) on Pybullet. This will enable anyone to 3D print and design their own version of \our.
In addition to hardware, we release an Isaac Gym and Pybullet-based simulator for \our. Its faithfulness to the real world is verified by performing sim2real in Sec.~\ref{section:inhand}. We release the sim2real platform to jumpstart lab research with \our.

\section{\our Applications}
\label{sec:learning_results}
First, we compare all of the hands in a grasping test with various everyday objects.  Next, we compare the two most robust, human-like hands, the 4-finger \our and the Allegro Hand \cite{allegro} against each other in a variety of machine learning tasks.  Teleoperation from human video demonstrates grasping capabilities and human-like form factor. Next, leveraging internet video shows the capability of learning from humans. Finally, we show \our on in-hand manipulation via sim2real, which demonstrates that the simulation and hardware are precise. In this task, \our is able to rotate the cube faster and is more robust to disturbances.  Please see the supplemental and our \href{https://leap-hand.github.io/}{website}  for videos of these results.

\begin{table}[!b]
\begin{center}
\resizebox{\linewidth}{!}{
\begin{tabular}{ll|cc| cc} 
 \toprule
  & & \multicolumn{2}{c|}{Success Rate} & \multicolumn{2}{c}{Completion Time (in \textit{s})}\\
 \# & Teleoperated Task & \our & Allegro & \our & Allegro\\   
 \midrule
 1 & Pickup Dice Toy & \textbf{1.0} & 0.9 & \textbf{6.5 (1.7)} & 8.6 (2.65) \\  

 2 & Pickup Dino Doll & \textbf{1.0} & 0.9 & \textbf{6.0 (1.5)} & 8.2 (3.49) \\ 

 3 & Box Rotation & \textbf{0.7} & 0.6 & \textbf{28.2 (15.7)} & 37.2 (12.6)\\ 

 4 & Scissor Pickup & 0.6 & \textbf{0.7} & 32.4 (7.8) & \textbf{28.6 (9.4)}\\ 

 5 & Cup Stack & \textbf{0.8} & 0.6 & \textbf{15.4 (7.0)} & 21.5 (7.6) \\  

 6 & Two Cup Stacking & \textbf{0.6} & 0.3 & \textbf{18.2 (9.2)} & 27.3 (11.0)\\  

 7 & Pour Cubes in Plate & \textbf{0.8} & 0.7 & \textbf{30.2 (15.2)} & 36.8 (17.7)\\  

 8 & Cup Into Plate & \textbf{0.8} & \textbf{0.8} & \textbf{6.2 (2.5)} & 10.6 (4.4) \\  

 9 & Open Drawer & \textbf{0.9} & \textbf{0.9} & \textbf{18.2 (11.2)} & 23.6 (12.3)\\ 

 10 & Open Drawer \& Pick & \textbf{0.7} & 0.6 & 37.2 (10.2) & \textbf{33.7 (8.1)} \\ 
 \hline \\ [-1.7ex]
 & \textbf{Outperform rate} & \textbf{9/10} & 3/10 & \textbf{8/10} & 2/10  \\
  \bottomrule
\end{tabular}}
\vspace{0.05in}
\caption{\small \textbf{Teleoperation---comparing \our and Allegro.} Success rate and average completion time of a trained operator completing a variety of teleoperated tasks. \our outperforms or matches the Allegro performance on 9/10 tasks.}
\label{tab:teleop}
\end{center}
\vspace{-0.1in}
\end{table}

\begin{figure}[t]
 \centering
 \includegraphics[width=\linewidth]{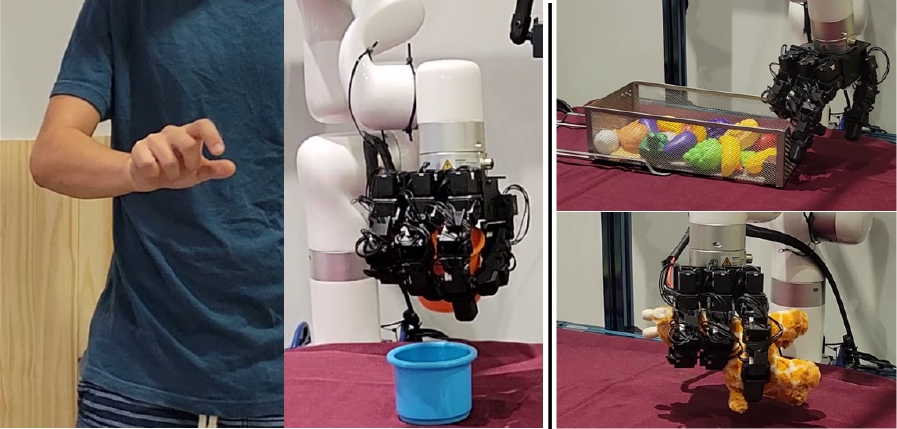}
 \vspace{-0.15in}
\caption{\small \textbf{Teleoperation and Behavior Cloning.} \textit{Left}: We perform dexterous teloperation using Telekinesis \cite{robo-telekinesis} with a single view color camera.  \textit{Right}: We perform behavior cloning from internet video and teeloperated demonstrations using Videodex \cite{videodex}.}
\vspace{-0.2in}
 \label{fig:telekinesis}
\end{figure}

\begin{table}[!t]
\centering
\resizebox{\linewidth}{!}{
\setlength{\tabcolsep}{2pt}
\begin{tabular}{lcccccc}
\toprule
Object & Grasp Type \cite{liu2014taxonomy} & LEAP & LEAP-C & Allegro & D'Manus & Inmoov \\
\midrule
\multicolumn{2}{l}{\textit{Power}:}\vspace{0.1em}\\
Mustard & Med. Palm+Pad & 20 & 20 & 13 & 8 & Y \\
Toy Kick Ball & Lrg. Palm+Pad & 20 & 20 & 9 & 20 & N \\
Golf Ball & Small Pad & 16 & 20 & 7 & 0 & Y  \\
Softball & Large Pad & 20 & 20 & 10 & 15 & N  \\
Drill & Trigger Press & 20 & 20 & 15 & 0 & N \\
Pringle Can & Power Palm & 19 & 20 & 20 & 0 & Y \\
Pan (from rim) & Disk Grasp & 20 & 20 & 20 & 14 & N\\
\midrule
\multicolumn{2}{l}{\textit{Intermediate}:}\vspace{0.4em}\\
Chopsticks & Tripod Grasp & 16 & 13 & 0 & 0 & N\\
Wood Cylinder & Cigarette Grasp & 4 & 5 & 0 & 0 & N \\
\midrule
\multicolumn{2}{l}{\textit{Precision}:}\vspace{0.4em}\\
1$"$ Cube & 2 Finger Precision & 20 & 20 & 20 & 0 & N \\ 
M\&M & Tip Pinch Grasp & Y & Y & Y & N & N \\
Wine Glass & Flat Hand Cupping & 20 & 20 & 4 & 0 & N \\
Credit Card & Lateral Pinch & 20 & 20 & 8 & 0 & N \\
\bottomrule
\end{tabular}}
\vspace{0.05in}
\caption{\small We test each robot hand on a variety of objects and grasps types and see how much perturbation force they can resist (in newtons).  The dexterous morphology of \our as well as its strong motors enables the cigarette and flat-hand cupping grasps.}
\vspace{-0.2in}
\label{tab:grasping}
\end{table}

\begin{table*}[t]
\centering
\resizebox{\linewidth}{!}{
\footnotesize
\begin{tabular}{lcccccccccccccc|c}
\toprule
& \multicolumn{2}{c}{Pick} & \multicolumn{2}{c}{Rotate} & \multicolumn{2}{c}{Open} & \multicolumn{2}{c}{Cover}  & \multicolumn{2}{c}{Uncover}  & \multicolumn{2}{c}{Place}  & \multicolumn{2}{c|}{Push} & \textbf{Overall}\\ 
& train & test & train & test & train & test & train & test & train & test  & train & test  & train & test \\
\midrule
Allegro Hand & 0.81 & 0.75 & \textbf{0.89} & 0.69 & 0.90 & \textbf{0.80} & 0.78 & 0.67  & \textbf{1.00} & \textbf{0.90} & 0.90 & 0.70 & \textbf{1.00} & \textbf{1.00} & {6/14}\\
\midrule
\our  & \textbf{0.92} & \textbf{0.84} & \textbf{0.89} & \textbf{0.72} & \textbf{0.94} & 0.76 & \textbf{0.80} & \textbf{0.75} & 0.96 & \textbf{0.90} & \textbf{0.94} & \textbf{0.75} & \textbf{1.00} & \textbf{1.00} & \textbf{12/14}\\
\bottomrule
\end{tabular}}
\vspace{-0.05in}
\caption{\small \textbf{Learning from videos via VideoDex~\cite{videodex}.} 
Hand policies are pretrained on internet videos of humans and finetuned using minimal ($\approx100$) teleoperated demos. On this practical use case, \our performs better on 12 of 14 \{task\}$\times$\{train, test\} pairs.
}
 \vspace{-0.1in}
\label{tab:videodex}
\end{table*}

\subsection{Grasping Test using Teleoperation}
\label{sec:grasping}
We compare each of the hands and their ability to perform different types of grasp when holding objects.  To quickly experiment and find these poses, we use the Manus Meta VR glove \cite{manus} to accurately teleoperate the first three hands (see appendix for details).  Since D'Manus is not anthropomorphic enough to teleoperate from human motion, we manually control keyframes for it.   We show various types of grasps that each hand can perform, and the amount of perturbation force they can resist (up to 20N). Once the object is grasped we push on it with the force gauge until it slips or the force gauge crosses 20N.  Because InMoov is too fragile to teleoperate and apply perturbation forces to, we only test if it can grasp the object securely and report these results in the table.  

Table \ref{tab:grasping} shows \our can grasp all objects and can perform both many power and precision grasps.  While \our and \ourablation perform similarly, the latter has weaker grasps because its MCP side motors cannot be used to adjust the grasp.  Allegro's motors are significantly weaker which leads to objects like the golf ball or the soccer ball to easily slip out.  Additionally, because its kinematics lacks adduction/abduction in an extended position, it cannot perform the tripod grasp for the chopsticks, the cigarette grasp for the wooden cylinder, or the flat hand cupping grasp for the wine glass properly.  D'Manus can complete extremely strong power grasps on larger objects, but its lack of opposability and inability to provide resistive force on all 4 sides of the objects hurts its performance.  Due to its large size, it fails to grasp smaller objects.

\subsection{Teleoperation from Uncalibrated Human Video}
Teleoperation enables control of high DOF robots in real-time via human feedback.  This is also a useful method for collection demonstrations. Because the Allegro Hand's morphology does not have a human-like MCP joint we must borrow the human-to-robot re-targeting method from Robotic Telekinesis \cite{robo-telekinesis} (Fig.~\ref{fig:telekinesis} (left)), that manually defines key vectors between palms and fingertips on both robot $v_i^r$ and human hand $v_i^h$. These vectors define an energy function $E_\pi$ which minimizes the distance between human hand poses (parameterized by the tuple $(\beta_{h}, \theta_{h})$)  and the robot hand poses $q$ scaled by $c_i$:
\begin{equation}
    E_{\pi}( \ (\beta_{h}, \theta_{h}), \  q \ ) = \sum_{i=1}^{10} || v_i^h- (c_i \cdot v_i^r) ||_2 ^2
    \label{eq:energy}
\end{equation}
\cite{robo-telekinesis} trains an MLP $H_R(.)$ to implicitly minimize the energy function described in Equation \ref{eq:energy}.

Because \our includes similar joints to a human, we can directly map joint angles between the human and robot.  In table \ref{tab:teleop}, we observe that \our performs better than Allegro Hand on 9/10 of these teleoperated tasks. \our is easier to control due to its better morphology, accuracy, and responsiveness to hand input.  As users in \cite{robo-telekinesis} mention, it is difficult to teleoperate a robot hand with an energy function.  Additionally, the better opposability and strength of \our allows the operator to reliably grasp objects that are difficult to grasp on the Allegro Hand.  While Allegro Hand needed to take breaks to avoid overheating like in \cite{handa2022dextreme}, \our kept running without a degredation of performance.

\subsection{Behavior Cloning from Demonstrations}
Behavior cloning enables agents to learn a policy for a particular task given demonstrations. However, collecting demonstrations for behavior cloning is expensive. We utilize video from Epic-Kitchens~\cite{EPICKITCHENS} as pre-training for our policy using VideoDex \cite{videodex} along with NDP \cite{bahl2020ndp}, see Fig.~\ref{fig:telekinesis} (right).
We also only use demonstrations collected from prior work \cite{videodex} on Allegro Hand and map those to \our. 
\our still outperforms the Allegro Hand in task performance as in Table \ref{tab:videodex}.  This is because of its consistency and strength while completing these tasks.

\begin{figure}[b!]
 \centering
 \vspace{-0.1in}
 \includegraphics[width=0.8\linewidth]{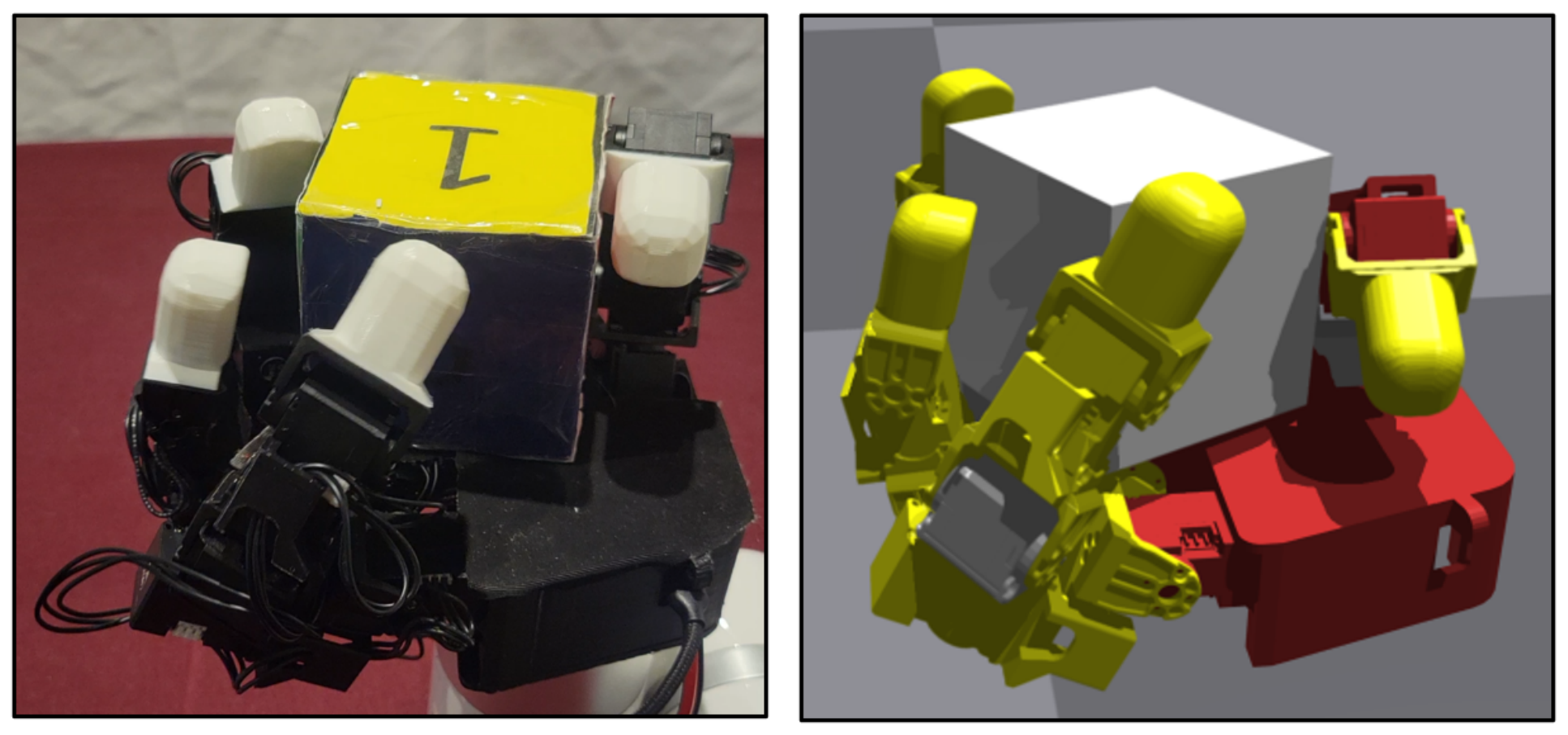}
 \vspace{-0.1in}
\caption{\small \textbf{Sim2Real transfer.} \textit{Left}: Simulated \our in Isaac Gym~\cite{makoviychuk2021isaac} completing an in-Hand manipulation task.  \textit{Right:} \our completing the same task in the real world. Please see our website ~\url{https://leap-hand.github.io/} for our open source pipeline.}
 \label{fig:inhand}
\end{figure}

\subsection{Sim2Real In-Hand Manipulation}
\label{section:inhand}
We perform in-hand rotation of a cube along an axis perpendicular to the palm.  \our can return current joint position, velocity, and torque and can be controlled from both torque or position commands.  In this case, the robot infers the object pose through the history of observed joint angles alone.  This is a challenging task since it is contact-rich, and the policy cannot directly observe the pose of the cube. The policy receives joint angles (16 values) from the motors and outputs the target joint angles (16) at 20 Hz which is passed as position commands to the motors.

We choose a GRU \cite{chung2014empirical} architecture for our policy. We first generate a cache of stable grasps similar to~\cite{qi2022hand}. The policy is then rewarded for turning the cube $r_\textrm{rot} = \textrm{clip}(\omega_z, -0.25, 0.25)$, where $\omega_z$ is the angular velocity along the vertical axis. We add additional penalties for deviation from the stable grasp pose, mechanical work done, motor torques, and object linear velocity. The scale for the rotation reward is 1.25. The scales for the penalties are -0.1, -1, -0.1, -0.3 respectively. We train PPO \cite{schulman2017proximal} with BPPT \cite{bppt} in IsaacGym \cite{makoviychuk2021isaac}. 

In simulation, we compare the average angular velocity of a cube for different hands and find that \our leads to faster rotations (Tab.~\ref{tab:sim_comparison}) than Allegro. This is because the joint structure of \our allows it to support the cube from the sides, whereas since Allegro does not have adduction/abduction it must let go of the cube periodically in order to re-orient it.

\begin{table}[b!]
\centering
\vspace{-0.1in}
\setlength{\tabcolsep}{3pt}\small
    \begin{tabular}{lc}
    \toprule
        Hand & Angular velocity (rad/s) \\
     \midrule
        Allegro & 0.0828 \\
        \ourablation & 0.2205 \\
        \our (ours) & \textbf{0.2288} \\
        \bottomrule
    \end{tabular}
    \caption{\small Comparison of angular velocity for the blind in-hand rotation of a cube in simulation. Since the Allegro Hand lacks abduction/adduction at its extended position, it has low angular velocity. However, \our and \ourablation have this ability and have better performance.}
    \label{tab:sim_comparison}
    \vspace{-0.1in}
\end{table}

\section{Conclusion and Future Work}
We introduce \our and its core design principles. Following these principles, we demonstrate that \our can perform exceedingly well compared to other hands on the market in strength, grasping, and durability.  We show its usefulness in a variety of real-world tasks, including teleoperation, behavior cloning, and sim2real.  We \textbf{open source} the URDF model, 3D CAD files, and a development platform with useful APIs. In future work, we plan to develop and integrate \our with low-cost touch sensors.

\vspace{0.1in}
\noindent\textbf{Acknowledgement}$\quad$
We thank Shikhar Bahl, Russell Mendonca, Unnat Jain and Jianren Wang for fruitful discussions about the project. KS is supported by NSF Graduate Research Fellowship under Grant No. DGE2140739. This work is supported by ONR N00014-22-1-2096 and the DARPA Machine Common Sense grant.

\bibliographystyle{IEEEtran}
\bibliography{IEEEabrv,main}

\end{document}